\documentclass[twocolumn]{article}

\usepackage[width=17cm,height=22cm]{geometry}
\usepackage[utf8]{inputenc}
\usepackage{fancyvrb}
\usepackage{authblk}

\usepackage{url}
\usepackage{amsmath}
\usepackage{amssymb}
\usepackage[dvips]{graphicx}
\usepackage{algorithm}
\usepackage[noend]{algorithmic}

\newcommand{\ea}[1]{#1 \emph{et al.}}
\newcommand{\set}[1]{\mathbb{#1}}
\newcommand{\algo}[1]{{\sc #1}}

\newtheorem{defin}{Definition}

\title{A feature construction framework based on outlier detection and discriminative pattern mining}

\author{Albrecht Zimmermann\\albrecht.zimmermann@insa-lyon.fr}
\affil{INSA Lyon, France}

\begin{document}

\maketitle

\begin{abstract}
No matter the expressive power and sophistication of supervised learning algorithms, their effectiveness is restricted by the features describing
the data.
This is not a new insight in ML and many methods for feature selection, transformation, and construction have been developed. But while this is on-going
for general techniques for feature selection and transformation, i.e. dimensionality \emph{reduction}, work on feature construction, i.e. \emph{enriching} the data, is by now mainly the domain of image, particularly character, recognition, and NLP.

In this work, we propose a new general framework for feature construction. The need for feature construction in a data set is indicated by \emph{class outliers} and discriminative pattern mining used to derive features on their $k$-neighborhoods.
We instantiate the framework with \algo{LOF} and \algo{C4.5-Rules}, and evaluate the usefulness of the derived features on a diverse collection of UCI data sets. The derived features are more often useful than ones derived by \algo{DC-Fringe}, and our approach is much less likely to overfit. But while a weak learner, \algo{Naive Bayes}, benefits strongly from the feature construction, the effect is less pronounced for \algo{C4.5}, and almost vanishes for an SVM leaner.

%
%

\end{abstract}

\section{Introduction}

Supervised learning -- concept, classifier, and regression learning -- is a core Machine Learning topic 
and decades of research have resulted in lots of algorithms. No matter the expressiveness or sophistication of a technique, however, its effectiveness is restricted by the data, and its representation.

As a result, researchers have also worked from early on to create useful representations. The resulting techniques can be grouped into three subfields:
\begin{enumerate}
 \item \emph{Feature selection} \cite{DBLP:conf/icdm/MolinaBN02,DBLP:journals/jmlr/GuyonE03,Liris-5964,Liris-6134} deals with the question of \emph{removing} irrelevant or redundant features from the data representation. Its goals are alleviating the ''curse of dimensionality`` -- the phenomenon that high-dimensional descriptor data points all seem to be equally similar to each other -- reducing running times of learning techniques, and preventing over-fitting.
 \item \emph{Feature transformation} or \emph{dimensionality reduction} \cite{Fodor02asurvey} has the same goals but achieves them by mathematical transformations of the matrix representation of the data, replacing the original representation.
 \item \emph{Feature construction} or \emph{constructive induction} \cite{yang1991scheme,DBLP:conf/gecco/ShaftiP05}, finally, aims at combining the existing features into new ones, and enriching the data with them, with the goal of making harder problems easier to model and increasing accuracy.
\end{enumerate}
This subdivision is not as clear-cut as we make it appear here: \cite{DBLP:journals/jmlr/GuyonE03}, for example, also discusses feature transformation techniques, referring to them as ''feature construction``. What is striking, however, is that after a flurry of early work on symbolic feature construction \cite{DBLP:conf/ijcai/MatheusR89,Aha91incrementalconstructive,yang1991scheme,pazzani1998constructive}, recent works employ genetic algorithms for feature construction \cite{DBLP:journals/gpem/Krawiec02,DBLP:conf/gecco/ShaftiP05}, focusing on constructing features from raw data \cite{DBLP:journals/jaise/ShaftiHGP13}, and for the purpose of NLP \cite{DBLP:conf/ijcai/GabrilovichM05,dhanasekaran2012research} and image recognition \cite{yang2008survey}.

For a data miner, this is somewhat puzzling, given that much recent DM research has addressed the problem of mining useful features for classification from complex data via discriminative pattern mining \cite{simpler-patterns,DBLP:conf/pkdd/ZimmermannBR10,DBLP:journals/sadm/ThomaCGHKSSYYB10}. Hence the motivation for this work.
Our contribution is two-fold:
\begin{enumerate}
 \item We propose a general framework for feature construction, employing outlier detection methods for identifying hard-to-model instances, and discriminative pattern mining for deriving useful features from their neighborhoods. In particular is this framework \emph{independent} of the learning algorithm employed.
 \item We instantiate our framework with \algo{LOF} as outlier detection technique and \algo{C4.5-Rules} as discriminative pattern miner, and evaluate and compare the usefulness of the derived features on a diverse collection of UCI data sets, using three learners of different strength. We show that our model-independent approach is far less likely to overfit and generates useful features.
\end{enumerate}

The paper is structured as follows: in the following section, we illustrate the problem setting and briefly discuss the four aspects to feature construction \cite{DBLP:conf/ijcai/MatheusR89}. In Section \ref{related-work}, we discuss related work. In Section \ref{framework}, we introduce our framework, and show how it addresses the four feature construction aspects. In Section \ref{instantiation}, we describe the concrete instantiation used in the experiments and report on the experimental results in Section \ref{experiments}, before concluding in Section \ref{conclusion}.

\section{Problem Illustration and Aspects of Feature Construction\label{problem-statement}}

\begin{figure}
\begin{center}
\includegraphics[width=0.6\linewidth]{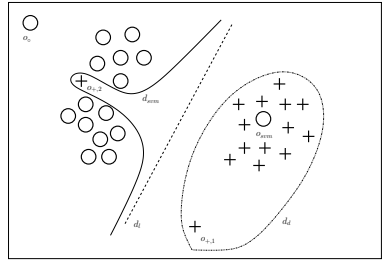}
\caption{Class outliers in a two-class problem\label{class-outliers}}
\end{center}
\end{figure}

As an illustration of the problem addressed in our work, consider Figure \ref{class-outliers}. Both classes are characterized by large clusters of typical instances, as well as several outlying points. There can be different reasons for this distribution: the training set could, e.g., \emph{not} be a representative sample -- a possible explanation for $o_{\circ}$ or $o_{+_1}$. An instance like $o_{+_2}$, however, either is mis-labeled or has incorrect attribute values (in other words, it is noise), or indicates that the description space is incomplete w.r.t. the underlying concepts.

A linear discriminative learner, inducing the decision surface $d_l$, or a concept learner characterizing the attribute-value space enclosed by $d_d$, will be unable to classify certain instances correctly. A more powerful learner, such as an SVM or an artificial neural network (ANN), could classify $o_{+,2}$ correctly but the representation would prevent it from separating $o_{svm}$ out. 

Finding a feature that takes the value $1$ for $o_{svm}$ and $0$ for its immediate neighborhood, as shown in Figure \ref{additional-dimension}, would make it possible to learn a decision surface for that instance. 
Hence, in this setting there is need for additional dimensions.

\subsection{Aspects of feature construction}

\begin{figure}
\centering
\includegraphics[width=0.4\linewidth]{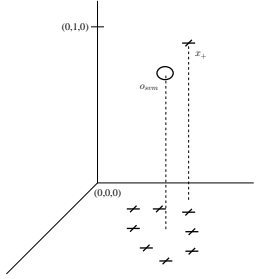}
\caption{Enriching data with an additional feature\label{additional-dimension}}
\end{figure}

Some data points are clearly not in need of augmentation: points belonging to the two clusters will be reliably modeled, and $o_{\circ}$ and $o_{+,1}$ can still be separated easily from the other class. There is also the question if a locally discriminating feature might not overfit the data too much and become useless outside of its local context (as in the case of $o_{+}$ in Figure \ref{additional-dimension}). The authors of \cite{DBLP:conf/ijcai/MatheusR89} turned this intuition systematic when they identified four aspects of feature construction:
\begin{quote}
\begin{enumerate}
 \item detection of when construction is required
 \item selection of constructors
 \item generalization of selected constructors
 \item evaluation of the new features
\end{enumerate}
\end{quote}
The first point is characterized by the authors writing: ``if the original feature set is sufficient for the [...] induction algorithm to acquire the target concept, feature construction is unnecessary.''. They identify three possible treatments: i) always perform construction, ii) analysis of the initial data set, iii) analysis of a model.

The second aspect has to do with the fact that the space of all attribute-value combinations can be too large to traverse, especially if arbitrary operators 
are allowed. They identify two approaches: i) initially limiting allowed operators, and ii) run time decisions that select one constructor over another based on algorithmic, data, concept, or domain knowledge biases.

The third aspect addresses over-fitting since new features might be highly specific to certain training instances. This includes removal of conjuncts, variable introduction etc.

The fourth aspect, finally, comes into play if the number of features becomes too large. The authors identify at least three approaches: i) keeping all features, ii) call on the user, or iii) order the features and keep the best ones.

\section{Related Work\label{related-work}}

Much early work on feature construction used decision trees as models, and based the newly constructed features on them. The setting for \algo{Fringe}, the technique developed in \cite{Pagallo89learningdnf}, is binary classification, with instances described by boolean attributes. The goal was the avoidance of the replication of partial paths in the decision tree. Detection consists of checking for the existence of positively-labeled leaf nodes at depth of at least two in the tree (model analysis), constructors are selected by choosing the tests in the last two decision nodes before the leaf (concept-based), which are combined using logical $AND$. The features are added to the representation, data instances are re-encoded, and the algorithm re-iterated until either no new features are generated or a pre-set number of iterations have been performed. The authors extended their approach to be able to deal with additional 
class 
labels, called \algo{Symmetric Fringe}. \ea{Matheus} \cite{DBLP:conf/ijcai/MatheusR89} introduced the four aspects of feature construction discussed in the preceding section, and extended \algo{Fringe} by considering tests in other regions of the tree, not just near the leaves. They keep the same detection and constructor selection, but add a generalization step -- in which constants can be replaced by variables -- and only keep the top-$27$ features, according to information gain. The resulting technique is called \algo{Citre} \cite{DBLP:conf/ijcai/MatheusR89} and allows the use of background knowledge in the selection, and evaluation steps. \algo{DC-Fringe} \cite{yang1991scheme}, finally, considers not only leaves but also the siblings of their parents. The authors claim that \algo{Citre} does not profit from using other tests than the ones at the fringe, and show that \algo{DC-Fringe} outperforms the earlier two techniques on a number of artificial data sets.

Other works from this period chose different classification models. Aha \cite{Aha91incrementalconstructive} used an incremental instance-based learner, using classification failure for detection. Classification is performed based on similarity to already encountered instances, and the new feature is chosen to be the conjunction of any pair of base attributes that maximizes the dissimilarity between the misclassified instance, and the differently labeled instances closest to it. They point towards the need for background knowledge for this approach to outperform \algo{Citre}. Pazzani \cite{pazzani1998constructive} proposed constructing features from the Cartesian product of existing attributes. Features are constructed from pairs of attributes, features can be deleted, only one such operation is performed per iteration, and the evaluation takes the form of accuracy estimation of a learner. In all these works, feature generation iterates repeatedly, using the same classification model.

At the same time, genetic algorithms were first used for feature construction to replace the local search for good features \cite{bensusan1996constructive}, and most recent work comes from this direction \cite{DBLP:conf/gecco/ShaftiP05,DBLP:journals/jaise/ShaftiHGP13}. These approaches allow for arbitrary combinations of atribute tests but pay with increased running times (especially since they typically iterate over a model inducer), which they address by limiting the number of base attributes used in feature construction.

The last reference also points to another trend in feature construction: recent works have mostly moved away from attribute-value settings and aim to construct features for text categorization \cite{DBLP:conf/ijcai/GabrilovichM05}, NLP \cite{dhanasekaran2012research}, and image recognition \cite{yang2008survey,rizoiu2013unsupervised}.

Somewhat separate from this are works in the data mining community, which aim to construct features useful for representing structured data, e.g. trees or graphs, for processing by machine learning techniques \cite{simpler-patterns,DBLP:conf/pkdd/ZimmermannBR10,DBLP:journals/sadm/ThomaCGHKSSYYB10}. The typical approach in those works consists of mining substructures that discriminate between two classes, and performing classification by SVM or decision tree learners.

\section{A Novel Feature Construction Framework\label{framework}}

Based on the problem illustration given in Section \ref{problem-statement}, we propose a novel, general framework for feature construction. We aim at a \emph{model independent} approach, i.e. one that can be run as a pre-processing step independent of the learner used. This also means that our approach does \emph{not} reiterate model learning. In the following, we lay out how we address the four aspects of feature construction.

\subsection{Detection -- Identifying class outliers}

We first need to address whether to construct features at all, and from which instances. For this purpose, we propose to use outlier detection techniques to identify atypical points in the data. Identifying outliers in the full data will not be useful since a point that appears as an outlier from the perspective of a cluster of differently-labeled instances could be easy to model for a classifier learner. Hence, we identify outliers from the perspective of \emph{individual} classes.
Assuming a data set ${\cal D}$, an \emph{outlier oracle} $OO(d,{\cal D}') \mapsto \{\text{true},\text{false}\}$ that can decide whether a instance $d$ is an outlier w.r.t. ${\cal D}'$, we can define the set of class outliers:
\begin{defin}
Given a set of labeled data ${\cal D} = \{(\vec{x},y)\}$, the set of outliers of class ${\cal D}_c = \{d = (\vec{x},c) \in {\cal D}\}$ is defined as ${\cal O}_{c,{\cal D}} = \{d \in {\cal D}_{c} \mid OO(d,{\cal D}_{c}) = \text{true}\}$. The union over all the outliers of all classes ${\cal O}_{C,{\cal D}} = \bigcup_{c} {\cal O}_{c,{\cal D}}$ is called the set of \emph{class outliers}.
\end{defin}
While we use the class label information in the detection step, we are \emph{not} making use of a particular classification model to detect the need for feature construction.
This has the advantage that our approach can be used as a pre-processing step for arbitrary classifier learners, but the potential disadvantage that we overestimate the need for feature construction.\footnote{Note that this outlier detection is not limited to the space of attribute-valued data. Any instance space in which a distance measure between instances is defined allows for the identification of outliers.}

\subsection{Constructor selection -- Assembling outliers' $k$-neighborhoods}

We intend to perform feature construction itself by mining discriminative patterns. The choice of pattern miner will impose some restrictions on the form new features can take, depending on the pattern language ${\cal L}_p$ employed.\footnote{Again, this step is not limited to the space of attribute-valued data. Any instance space in which discrimative patterns can be mined can be worked on.}
In addition to this, we employ a data-biased selection mechanism, collecting the $k$-neighborhoods of class outliers: 
\begin{defin}
Given an instance $d$ and a distance measure $\delta: {\cal D} \times {\cal D} \mapsto \mathbb{R}$, its $k$-\emph{neighborhood} is defined as: ${\cal N}_{k}(d) = \{d' \in {\cal D}\setminus\{d\} \text{ s.t. } |\{d'' \mid \delta(d'',d) \leq \delta(d',d)\}| \leq k-1\}$
\end{defin}
and, if they belong to different classes, mining discriminative patterns from them. This condition -- that different classes have to be present -- can be considered part of the detection step: class outliers like $o_{\circ}$ in Figure \ref{class-outliers} do not indicate a need for feature construction.

This data selection step reduces the space of possible constructors to those present in this $k$-neighborhood, making the feature construction step tractable. Additionally, they can be expected to be relatively small, aiding further in tractability.

\subsection{Generalization -- Merging $k$-neighborhoods}

According to \cite{DBLP:conf/ijcai/MatheusR89}, generalization of features occurs \emph{after} they have been mined. In contrast to this, we address the generalization aspect \emph{before} feature construction by comparing the $k$-neighborhoods of all pairs of class outliers, and merging them if at least $50\%$ of the instances of the smaller neighborhood are included in the larger one. This is transitive: if an outlier's neighborhood has at least half its instances contained in a second neighborhood, those instances will also be contained in any neighborhood the second one was merged with.

\subsection{Evaluation -- Removing inconsistent and insufficiently supported features}

Since we intend our feature construction approach to be used in a pre-processing step, we do not use any model's classification accuracy to evaluate features during construction. Since we \emph{do} intend to mine features that discriminate outliers from their differently-labeled neighbors, however, we only select \emph{consistent} patterns, i.e. ones that cover instances of only a single class within a $k$-neighborhood: 
\begin{defin}
Given a matching function \emph{match}$: {\cal L}_p \times {\cal D} \mapsto \{0,1\}$, the \emph{cover} of a pattern $\pi \in {\cal L}_p$ in a data set ${\cal D}'$ is defined as $\text{cov}(\pi,{\cal D}') = \{d \in {\cal D}' \mid \text{match}(\pi,d) = 1\}$.
\end{defin}

\begin{defin}
Given a pattern $\pi$, $k$-neighborhood ${\cal N}_k$ it has been mined from, we call a pattern \emph{consistent} iff $\forall d_i = (\vec{x}_i,y_i), d_j = (\vec{x_j},y_j) \in \text{cov}(\pi, {\cal N}_k): y_i = y_j$.
\end{defin}

Since in that case features could overfit, we also impose a minimum support constraint on a feature.

\begin{defin}
Given a pattern $\pi$, its \emph{support} on a data set ${\cal D}'$ is defined as $\text{supp}(\pi, {\cal D}') = |\text{cov}(\pi,{\cal D}')|$.
\end{defin}

Finally, if a feature appears several times, it is selected only once.

\subsection{Algorithm}

The meta-algorithm describing the workings of our approach is given as Algorithm \ref{meta-algorithm}. We will refer to our framework as \algo{CobFC} -- for class-outlier based feature construction -- in the rest of the paper.

\begin{algorithm}
\begin{algorithmic}[1]
\caption{The \algo{CobFC} algorithm\label{meta-algorithm}}
\STATE Given: data set $\cal D$, set of class labels $C$, support threshold $\theta$
\STATE Return: set of consistent discriminative patterns $\set{F}$
\STATE
\STATE ${\cal O}_{C,{\cal D}} = \emptyset$
\FOR[Collecting class outliers]{$c \in C$}
\STATE ${\cal O}_{C,{\cal D}} = {\cal O}_{C,{\cal D}} \cup {\cal O}_{c,{\cal D}}$
\ENDFOR
\STATE $\set{N} = \emptyset$
\FOR{$o \in {\cal O}_{C,{\cal D}}$}
\IF[Different classes in the neighborhood]{$\exists d_i=(\vec{x}_i,y_i),d_j=(\vec{x}_j,y_j) \in {\cal N}_k(o): y_i \neq y_j$}
\STATE $\set{N} = \set{N} \cup \{{\cal N}_k(o)\}$
\ENDIF
\ENDFOR
\WHILE[Merging overlapping neighborhoods]{$\exists {\cal N}_k(o),{\cal N}_k(o') \in \set{N}: (\frac{|{\cal N}_k(o') \cap {\cal N}_k(o)|}{\min\{|{\cal N}_k(o')|,|{\cal N}_k(o)|\}} \geq 0.5)$}
\STATE $\set{N} = ((\set{N} \cup \{{\cal N}_k(o') \cup {\cal N}_k(o)\}) \setminus {\cal N}_k(o')) \setminus {\cal N}_k(o)$
\ENDWHILE
\FOR[Mining discrimative patterns]{${\cal N}_k(o) \in \set{N}$}
\FORALL{$f \in \text{PM}({\cal N}_k(o))$}
\IF{$f \text{ is consistent} \wedge \text{supp}(f, {\cal D}) \geq \theta$}
\STATE $\set{F} = \set{F} \cup \{f\}$
\ENDIF
\ENDFOR
\ENDFOR
\RETURN{$\set{F}$}
\end{algorithmic}
\end{algorithm}

\section{A Framework Instantiation\label{instantiation}}

We aim to give an impression of our framework's performance compared to other approaches, and will use a proof-of-concept instantation of \algo{CobFC}.

In the experimental study, we work in the usual space of data described by a vector of attribute values: given a set of attributes ${\cal A} = \{A_1,\ldots,A_d\}$, having domains $dom(A_i)$, each instance $d$ is a tuple $(\vec{x},y)$ with $\vec{x} = \langle x_1,\ldots,x_d\rangle, x_i \in dom(A_i)$, $y\in dom(C) = \{c_1,\ldots,c_m\}$ a class label.

As outlier oracle, we choose the \algo{LOF} algorithm \cite{DBLP:conf/sigmod/BreunigKNS00} in the version available for download by its authors as part of the ELKI package.\footnote{At \url{http://www.dbs.ifi.lmu.de/research/KDD/ELKI/}.}

To mine discriminative patterns, we use the C4.5 implementation of WEKA \cite{weka} (J48), with pruning turned off and minimum number of instances set to $1$. We transform the resulting tree into conjunctive rules and use the left-hand side of these rules as features for augmentation. This results in binary features.
Notably, this means that we can use two off-the-shelf solutions for instantiating our framework and do not have to write tailor-made code, making our approach highly modular. The ELKI package offers numerous outlier detection methods, which we plan to explore in more detail in the future.

\section{Experimental Setup\label{experiments}}

We expect the \algo{CobFC} features to improve the modeling of classification problems and therefore the estimated classification accuracy. To evaluate classification performance, we perform a ten-fold cross-validation on a number of UCI data sets \cite{uci}.
We aimed for data sets of different dimensionality, size, and distribution and number of classes to allow us to evaluate the behavior of class outlier detection and the effects of derived features thoroughly.\footnote{We do not list their characteristics. Those can be found in supplementary material at \url{http://www.scientific-data-mining.org/supplementary-material.html} or accessed at the UCI repository, however.}
For each fold, class outliers and their $k$-neighborhoods are extracted from the combined training folds, and conjunctive patterns mined on these subsets. To evaluate the effect on learners of different strength, we used LibSVM, as well as the J48 decision tree and the Na{\"i}ve Bayes classifier  implementations contained in the WEKA workbench \cite{weka}.

\subsection{Comparison techniques}

We compare our approach to the feature construction technique \algo{DC-Fringe} \cite{yang1991scheme}, which had been shown to outperform the other decision tree based feature construction methods and alternatives using other models. That method shows some differences to our approach:
\begin{itemize}
 \item Detection is handled by inspecting leaf nodes of depth at least two in a learned decision tree. As in our approach, this will lead to a focus on small subsets that lie near decision boundaries but whether different classes are present in such leaves has no effect.
 \item Individual features are constructed as conjunction or disjunction of the last two tests leading to a given leaf. For details, we refer the reader to the original publication. Thus, while restricted in size, features are constructed using more powerful operators.
 \item The process is reiterated, learning a new decision tree on the augmented data, until a stopping criterion is reached, in contrast to our method, which constructs features only once.
\end{itemize}

\algo{DC-Fringe} was proposed in the context of concept learning and binary attributes. While most of the method can be transferred to a multi-class setting, and multi-valued or numerical attributes in a straight-forward manner, the stopping criterion cannot. \algo{DC-Fringe} stops when there are no more leaves having at least depth two but we noticed in our experiments that this case is typically not reached. Instead, decision trees stabilize, having the same form, and therefore leading to the same features, iteration after iteration. Hence, we modify the stopping criterion such that feature construction stops when the decision tree does not change from one iteration to the next. We used the J48 decision tree learner.

As a baseline comparison, we chose a method suggested by an anonymous reviewer of an earlier version of this work: we remove the class outliers from the training data, and train on the reduced data set. While this is somewhat counter to our purpose, it might prevent over-fitting effects caused by attempts to model unaugmented class outliers and therefore improve testing accuracies.

\subsection{Experimental results outlier detection and feature generation\label{feature-generation}}

We used Euclidean distance for identifying outliers in data sets with numerical attributes, normalizing attribute values. We used Manhattan distance for identifying outliers in data sets with nominal and ordinal attributes, binarizing attributes to that end. 

\begin{table*}
\scriptsize
\centering
\begin{tabular}{l|c|c|c|c|c||c|c}
& \multicolumn{3}{c|}{Outliers} & Merged & Number of & \multicolumn{2}{c}{\algo{DC-Fringe}}\\
Data Set & Avg. \# & Classes w/o & Folds w/o & neighborhoods & features & Features & Iterations\\\hline
Breast Cancer (Wisc.) & 44.30 & 0.00 & 0 & 30.40 & 64.50 & 255.00 & 15.80\\
Diabetes (Pima)& 47.60 & 0.00 & 0 & 40.60 & 119.20 & 2572.90 & 47.10\\
Ecoli & 20.50 & 1.00 & 0 & 15.40 & 26.10 & 506.50 & 16.80\\
Glass & 35.30 & 1.30 & 0 & 28.30 & 56.00 & 418.40 & 21.40\\
Heart Statlog & 17.50 & 0.00 & 0 & 16.40 & 56.60 & 307.10 & 15.30\\
Ionosphere & 27.30 & 0.00 & 0 & 19.00 & 23.80 & 81.00 & 8.20\\
Iris & 8.90 & 0.00 & 0 & 6.30 & 7.00 & 18.70 & 5.20\\
Liver Disorders & 27.50 & 0.00 & 0 & 14.40 & 60.50 & 1045.40 & 33.30\\
Optdigits & 220.80 & 0.00 & 0 & 203.80 & 246.60 & 6463.00 & 25.80\\
Page blocks & 528.40 & 0.00 & 0 & 244.60 & 329.70 & 1247.00 & 16.56\\
Pendigits & 757.40 & 0.00 & 0 & 587.90 & 414.20 & 5035.10 & 22.60\\
Segment & 163.60 & 0.00 & 0 & 79.60 & 149.90 & 538.30 & 12.20\\
Segnoise & 105.40 & 0.00 & 0 & 98.00 & 410.60 & 349.70 & 10.40\\
Sonar & 7.80 & 0.00 & 0 & 7.30 & 15.10 & 108.30 & 10.90\\
Spambase & 354.20 & 0.00 & 0 & 126.60 & 408.40 & 8102.50 & 66.50\\
Spectrometer & 39.80 & 8.30 & 0 & 33.70 & 126.50 & 3585.30 & 32.20\\
Vehicle & 43.00 & 0.00 & 0 & 28.70 & 111.30 & 2971.80 & 23.30\\ \hline \hline
Audiology & 16.70 & 4.20 & 0 & 10.80 & 31.50 & 350.10 & 12.00\\
Breast Cancer & 20.30 & 0.00 & 0 & 18.90 & 85.10 & 854.60 & 22.40\\
Car & 247.90 & 1.00 & 0 & 246.70 & 410.00 & 236.80 & 13.90\\
Dermatology & 20.70 & 0.40 & 0 & 19.70 & 35.20 & 84.80 & 9.70\\
Kr-vs-kp (Chess)& 184.70 & 0.00 & 0 & 27.60 & 50.50 & 370.10 & 13.50\\
Lung Cancer & 1.22 & 2.22 & 1 & 1.22 & 2.89 & 44.80 & 8.70\\
Lymph & 9.30 & 0.60 & 0 & 8.70 & 26.60 & 138.70 & 9.10\\
Mol. Biol. (Prom.) & 47.70 & 1.00 & 0 & 37.20 & 40.40 & 30.60 & 6.10\\
Nursery & 904.90 & 0.20 & 0 & 904.90 & 2500.60 & 423.80 & 14.00\\
Postop. Patient Data & 6.50 & 0.00 & 0 & 6.10 & 17.50 & 171.00 & 10.70\\
Primary Tumor & 20.20 & 5.80 & 0 & 18.90 & 80.70 & 1509.60 & 19.30\\
Soybean & 34.10 & 6.70 & 0 & 29.20 & 56.50 & 495.70 & 13.40\\
Splice & 97.10 & 1.00 & 0 & 75.00 & 126.10 & 1659.22 & 20.00\\
Tic-tac-toe & 62.50 & 0.00 & 0 & 62.50 & 168.20 & 317.60 & 12.10\\
Voting Record & 47.40 & 0.00 & 0 & 25.50 & 28.80 & 105.80 & 8.00\\
\end{tabular}
\caption{Quantitative characteristics of the feature construction step on a selection of UCI data sets\label{quantitative-characteristics}}
\end{table*}

In this first section of the experimental evaluation, we report on the quantitative characteristics of the feature construction step: number of class outliers identified, number of merged $k$-neighborhoods, and number of features derived.

As Table \ref{quantitative-characteristics} shows, there is a significant number of class outliers for most data sets (column 2) but in some data sets some classes do not have outliers at all (column 3), often because they are too small for the concept of class outliers to have meaning. Depending on the data set, merging outliers' $k$-neighborhoods can reduce their number by half (e.g. for the \emph{Spambase} data set), or leave them unchanged (\emph{Nursery}). The number of features, finally, can range from below ten to thousands.

It can also be seen that \algo{DC-Fringe} constructs far more features (with the notable exception of \emph{Nursery}), and performs multiple iterations, each of which will be costlier due to the higher dimensionality of the data.

\subsection{Experimental evaluation: Classification accuracy\label{feature-evaluation}}


We chose the RBF kernel for LibSVM, selecting the $\lambda$ and $C$-parameters by grid search via internal five-fold cross-validation from the ranges $[2^{-15},2^3]$, and $[2^{-5},2^{15}]$, respectively, doubling the value in each step. 
J48 and Na{\"i}ve Bayes were run using the standard settings predefined in WEKA. 
We want to stress again that these are proof-of-concept experiments, and we did not fine-tune the aspects of our framework. But, anticipating reviewer comments, we have used the Friedman-Nemenyi procedure as stipulated in \cite{garcia2008extension}, finding that none of the three techniques leads to significantly better results than any other.

 \begin{table*}[ht]
 \scriptsize
 \centering
 \begin{tabular}{l|c|c|c|c|c}
 & \multicolumn{2}{c|}{Original} & \algo{CobFC} & DC-Fringe & Baseline\\
 Data set & Train & Test & Test & Test & Test\\\hline
 Breast Cancer (Wisc.) & $96.11 \pm 0.26$ & $96.14 \pm 2.24$& $\mathbf{96.43 \pm 2.80}$ & $\underline{96.00 \pm 2.59}$ & $\underline{95.43 \pm 3.21}$ \\ 
Diabetes (Pima) & $76.27 \pm 0.73$ & $76.18 \pm 4.83$& $74.62 \pm 4.52$ & $\underline{70.83 \pm 3.20}$ & $\underline{74.62 \pm 5.92}$ \\ 
Ecoli & $88.33 \pm 1.43$ & $85.43 \pm 3.82$& $83.65 \pm 6.25$ & $\underline{78.24 \pm 7.09}$ & $\underline{84.53 \pm 4.76}$ \\ 
Glass & $54.77 \pm 2.79$ & $47.36 \pm 11.52$& $\mathbf{65.06 \pm 12.77}$ & $\mathbf{72.94 \pm 10.37}$ & $40.28 \pm 7.83$ \\ 
Heart Statlog & $85.76 \pm 1.12$ & $83.70 \pm 7.45$& $\mathbf{84.07 \pm 8.74}$ & $\underline{81.85 \pm 6.86}$ & $83.33 \pm 7.04$ \\ 
Ionosphere & $83.57 \pm 0.86$ & $82.63 \pm 5.76$& $\mathbf{87.75 \pm 3.80}$ & $\mathbf{89.46 \pm 4.67}$ & $\mathbf{84.90 \pm 5.86}$ \\ 
Iris & $95.85 \pm 0.87$ & $96.00 \pm 4.66$& $91.33 \pm 6.32$ & $\underline{94.67 \pm 5.26}$ & $\underline{96.00 \pm 6.44}$ \\ 
Liver Disorders & $56.88 \pm 2.75$ & $53.34 \pm 5.82$& $\mathbf{63.18 \pm 6.60}$ & $\mathbf{66.99 \pm 6.50}$ & $\mathbf{63.53 \pm 5.46}$ \\ 
Optdigits & $91.82 \pm 0.16$ & $91.30 \pm 1.03$& $\mathbf{92.17 \pm 1.32}$ & $\mathbf{93.10 \pm 1.25}$ & $\underline{90.87 \pm 1.07}$ \\ 
Pendigits & $85.86 \pm 0.20$ & $85.72 \pm 0.92$& $84.42 \pm 1.47$ & $\mathbf{88.24 \pm 1.08}$ & $\underline{85.32 \pm 1.10}$ \\ 
Spambase & $79.51 \pm 0.17$ & $81.17 \pm 1.17$& $\mathbf{91.64 \pm 0.75}$ & $\mathbf{93.61 \pm 0.82}$ & $\underline{78.85 \pm 1.80}$ \\ 
Page blocks & $90.31 \pm 0.92$ & $90.28 \pm 1.77$& $90.01 \pm 1.90$ & $89.56 \pm 2.32$ & $\underline{87.96 \pm 5.93}$ \\ 
Segment & $80.49 \pm 0.24$ & $80.17 \pm 2.23$& $\mathbf{89.70 \pm 1.82}$ & $\mathbf{91.69 \pm 1.55}$ & $\mathbf{83.03 \pm 1.08}$ \\ 
Segnoise & $81.88 \pm 0.30$ & $80.87 \pm 1.78$& $\mathbf{89.20 \pm 3.20}$ & $\mathbf{89.20 \pm 2.45}$ & $\underline{80.33 \pm 3.27}$ \\ 
Sonar & $72.28 \pm 1.35$ & $67.83 \pm 7.99$& $\mathbf{69.26 \pm 6.73}$ & $\mathbf{77.45 \pm 10.97}$ & $65.95 \pm 8.65$ \\ 
Spectrometer & $62.08 \pm 1.08$ & $42.74 \pm 5.63$& $\mathbf{43.30 \pm 5.91}$ & $\underline{42.55 \pm 4.51}$ & $\underline{41.61 \pm 7.48}$ \\ 
Vehicle & $46.93 \pm 0.76$ & $45.27 \pm 2.26$& $\mathbf{49.75 \pm 3.86}$ & $\mathbf{52.01 \pm 4.11}$ & $\mathbf{49.63 \pm 3.39}$ \\\hline\hline 
 Audiology & $93.76 \pm 0.90$ & $75.24 \pm 5.85$& $\mathbf{76.52 \pm 7.77}$ & $74.82 \pm 6.06$ & $65.08 \pm 7.19$ \\ 
Breast Cancer & $74.98 \pm 1.27$ & $72.38 \pm 8.80$& $\underline{70.18 \pm 10.72}$ & $\underline{67.49 \pm 10.90}$ & $\underline{72.34 \pm 7.57}$ \\ 
Car & $88.22 \pm 0.26$ & $86.86 \pm 2.64$& $\mathbf{87.03 \pm 3.73}$ & $\mathbf{91.44 \pm 8.10}$ & $\underline{81.60 \pm 2.77}$ \\ 
Dermatology & $99.33 \pm 0.24$ & $98.92 \pm 1.40$& $97.82 \pm 2.79$ & $95.11 \pm 2.47$ & $98.38 \pm 2.61$ \\ 
Kr-vs-kp (Chess)& $85.79 \pm 0.37$ & $85.39 \pm 2.13$& $\mathbf{87.20 \pm 2.92}$ & $81.70 \pm 5.30$ & $\mathbf{87.58 \pm 2.66}$ \\ 
Lung Cancer & $91.31 \pm 2.50$ & $52.50 \pm 32.41$& $\underline{49.17 \pm 28.45}$ & $\underline{36.67 \pm 23.31}$ & $51.67 \pm 27.72$ \\ 
Lymph & $88.29 \pm 1.11$ & $82.57 \pm 9.93$& $\mathbf{83.14 \pm 6.38}$ & $\mathbf{84.48 \pm 8.30}$ & $\mathbf{83.81 \pm 9.03}$ \\ 
Mol. Biol. (Prom.) & $98.95 \pm 0.49$ & $91.36 \pm 7.34$& $91.27 \pm 8.51$ & $\underline{87.64 \pm 6.50}$ & $\underline{50.00 \pm 3.71}$ \\ 
Nursery & $90.40 \pm 0.13$ & $90.31 \pm 1.11$& $\mathbf{93.83 \pm 0.88}$ & $\mathbf{90.47 \pm 2.94}$ & $\mathbf{92.52 \pm 0.82}$ \\ 
Postop. Patient Data & $67.04 \pm 3.08$ & $54.44 \pm 12.23$& $\mathbf{56.67 \pm 13.30}$ & $\underline{53.33 \pm 17.21}$ & $\mathbf{58.89 \pm 15.76}$ \\ 
Primary Tumor & $56.11 \pm 0.83$ & $43.98 \pm 5.12$& $42.48 \pm 7.61$ & $38.95 \pm 5.41$ & $\mathbf{44.27 \pm 7.09}$ \\ 
Soybean & $93.80 \pm 0.41$ & $93.26 \pm 3.47$& $\mathbf{93.56 \pm 2.87}$ & $\underline{91.51 \pm 3.23}$ & $90.19 \pm 2.94$ \\ 
Splice & $96.12 \pm 0.19$ & $95.89 \pm 1.17$& $94.98 \pm 1.17$ & $92.07 \pm 1.74$ & $\underline{95.80 \pm 1.17}$ \\ 
Tic-tac-toe & $69.38 \pm 1.05$ & $67.74 \pm 3.90$& $\mathbf{68.48 \pm 5.16}$ & $\mathbf{89.35 \pm 2.72}$ & $\mathbf{68.37 \pm 3.94}$ \\ 
Voting Record & $91.55 \pm 0.65$ & $91.50 \pm 4.16$& $\mathbf{94.25 \pm 3.45}$ & $\underline{90.33 \pm 5.06}$ & $90.58 \pm 4.73$ \\\hline 
\multicolumn{3}{c|}{Improvements} & 21 & 13 & 10\\
\multicolumn{3}{c|}{Over-fitting} & 2 & 12 & 14\\
  \end{tabular}
\caption{Results of Naive Bayes for data sets pre-processed with \algo{CobFC}, \algo{DC-Fringe}, and the baseline approach, respectively\label{nb-accuracies}}
 \end{table*}

An obvious risk with feature construction lies in over-fitting. We therefore report testing accuracies in the following way. In each table, for \algo{Naive Bayes}, \algo{SVM}, and \algo{J48}, we show training and testing accuracies on the original, unaugmented data, as well as testing accuracies for data that has been pre-processed, either having been augmented by features
 or having had the class outliers removed from the training data.
 
 If the \emph{testing} accuracy for a pre-processing approach is higher than for the unaugmented data, the value is shown in {\bf bold}, whereas if the \emph{training} accuracy is higher, but the testing accuracy equal or lower -- i.e., when we observe over-fitting -- the value is \underline{underlined}.\footnote{Full tables with training/testing accuracies can be downloaded at \url{http://www.scientific-data-mining.org/supplementary-material.html}.}

 \begin{table*}[ht]
 \scriptsize
 \centering
 \begin{tabular}{l|c|c|c|c|c}
 & \multicolumn{2}{c|}{Original} & \algo{CobFC} & DC-Fringe & Baseline\\
 Data set & Train & Test & Test & Test & Test\\\hline
 Breast Cancer (Wisc.) & $97.12 \pm 0.23$ & $96.57 \pm 1.80$& $\underline{96.43 \pm 1.81}$ & $\underline{95.57 \pm 2.37}$ & $\underline{95.14 \pm 1.80}$ \\ 
Diabetes & $77.79 \pm 1.29$ & $74.62 \pm 6.48$& $\underline{74.49 \pm 5.99}$ & $\underline{69.14 \pm 4.81}$ & $\mathbf{75.66 \pm 5.18}$ \\ 
Ecoli & $89.58 \pm 1.54$ & $87.82 \pm 5.41$& $84.22 \pm 4.30$ & $\underline{76.18 \pm 8.91}$ & $\underline{86.33 \pm 5.37}$ \\ 
Glass & $82.29 \pm 4.44$ & $70.15 \pm 7.41$& $\mathbf{75.30 \pm 8.52}$ & $\mathbf{73.42 \pm 8.29}$ & $\underline{68.70 \pm 8.95}$ \\ 
Heart Statlog & $86.42 \pm 1.21$ & $80.00 \pm 8.76$& $\mathbf{82.22 \pm 8.69}$ & $\underline{75.56 \pm 8.04}$ & $77.41 \pm 4.77$ \\ 
Ionosphere & $98.26 \pm 1.24$ & $93.74 \pm 3.75$& $\mathbf{95.16 \pm 3.82}$ & $\underline{91.74 \pm 4.14}$ & $92.02 \pm 3.51$ \\ 
Iris & $98.37 \pm 1.20$ & $96.67 \pm 3.51$& $93.33 \pm 6.29$ & $\underline{95.33 \pm 4.50}$ & $\underline{96.00 \pm 6.44}$ \\ 
Liver Disorders & $77.46 \pm 0.76$ & $73.08 \pm 8.01$& $\underline{72.24 \pm 9.29}$ & $\underline{64.36 \pm 7.68}$ & $69.32 \pm 7.73$ \\ 
Spambase & $95.83 \pm 0.97$ & $93.92 \pm 1.42$& $\underline{93.38 \pm 0.75}$ & $75.61 \pm 42.28$ & $91.52 \pm 2.20$ \\ 
Page Blocks & $98.21 \pm 0.12$ & $96.02 \pm 0.68$& $\underline{95.89 \pm 0.79}$ & $83.89 \pm 33.91$ & $\underline{94.90 \pm 0.83}$ \\ 
Segment & $99.32 \pm 0.22$ & $97.36 \pm 0.75$& $\mathbf{97.88 \pm 0.75}$ & $\mathbf{97.88 \pm 1.07}$ & $\underline{95.67 \pm 1.34}$ \\ 
Segnoise & $99.73 \pm 0.31$ & $96.33 \pm 1.14$& $\mathbf{96.80 \pm 1.21}$ & $\mathbf{97.20 \pm 0.98}$ & $\underline{95.40 \pm 1.39}$ \\ 
Sonar & $99.73 \pm 0.84$ & $86.57 \pm 6.25$& $79.76 \pm 8.20$ & $\underline{78.40 \pm 9.14}$ & $\underline{86.07 \pm 8.28}$ \\ 
Spectrometer & $10.36 \pm 0.11$ & $10.36 \pm 1.01$& $10.36 \pm 1.01$ & $10.36 \pm 1.01$ & $\underline{10.36 \pm 1.01}$ \\ 
Vehicle & $94.77 \pm 1.36$ & $83.45 \pm 3.55$& $\underline{83.09 \pm 2.51}$ & $\underline{80.84 \pm 2.83}$ & $\underline{82.14 \pm 2.64}$ \\ 
Audiology & $99.80 \pm 0.34$ & $78.72 \pm 5.98$& $76.94 \pm 7.55$ & $\mathbf{80.08 \pm 9.71}$ & $77.43 \pm 5.62$ \\\hline\hline 
Breast Cancer & $82.86 \pm 6.14$ & $72.38 \pm 7.11$& $\underline{72.02 \pm 6.39}$ & $\underline{66.10 \pm 8.56}$ & $\underline{71.66 \pm 5.94}$ \\ 
Car & $100.00 \pm 0.00$ & $99.71 \pm 0.41$& $99.25 \pm 0.72$ & $\mathbf{99.83 \pm 0.39}$ & $95.25 \pm 1.77$ \\ 
Dermatology & $99.24 \pm 0.43$ & $98.10 \pm 2.87$& $\underline{97.55 \pm 2.69}$ & $\underline{96.19 \pm 2.62}$ & $\underline{97.00 \pm 2.69}$ \\ 
Kr-vs-kp (Chess)& $99.98 \pm 0.02$ & $99.72 \pm 0.31$& $99.47 \pm 0.33$ & $\underline{99.62 \pm 0.29}$ & $99.19 \pm 0.49$ \\ 
Lung Cancer & $76.00 \pm 26.65$ & $40.00 \pm 19.56$& $36.67 \pm 17.21$ & $40.00 \pm 25.09$ & $\underline{36.67 \pm 17.21}$ \\ 
Lymph & $96.02 \pm 4.89$ & $82.57 \pm 8.88$& $\underline{81.05 \pm 8.36}$ & $\underline{79.71 \pm 10.97}$ & $\mathbf{84.52 \pm 8.27}$ \\ 
Mol. Biol. (Prom.) & $99.68 \pm 0.51$ & $90.36 \pm 6.68$& $\underline{90.36 \pm 7.94}$ & $\underline{88.55 \pm 7.56}$ & $\underline{50.00 \pm 3.71}$ \\ 
Postop. Patient Data & $71.11 \pm 0.64$ & $71.11 \pm 5.74$& $71.11 \pm 5.74$ & $\underline{56.67 \pm 22.50}$ & $\underline{67.78 \pm 11.05}$ \\ 
Primary Tumor & $63.68 \pm 4.50$ & $48.09 \pm 6.73$& $\mathbf{49.27 \pm 7.88}$ & $\underline{44.29 \pm 7.90}$ & $46.61 \pm 7.97$ \\ 
Soybean & $97.48 \pm 1.47$ & $93.12 \pm 2.86$& $\underline{93.12 \pm 3.60}$ & $\underline{91.80 \pm 3.93}$ & $\underline{92.53 \pm 3.42}$ \\ 
Splice & $98.74 \pm 1.24$ & $96.83 \pm 0.76$& $\underline{96.05 \pm 0.89}$ & $81.95 \pm 36.15$ & $\underline{96.36 \pm 0.81}$ \\ 
Tic-tac-toe & $100.00 \pm 0.00$ & $100.00 \pm 0.00$& $98.12 \pm 1.61$ & $97.91 \pm 1.77$ & $97.92 \pm 1.55$ \\ 
Voting & $97.93 \pm 0.66$ & $96.55 \pm 2.69$& $96.10 \pm 3.87$ & $\underline{94.01 \pm 3.30}$ & $\mathbf{97.25 \pm 2.80}$ \\ \hline
\multicolumn{3}{c|}{Improvements} & 6 & 5 & 3\\
\multicolumn{3}{c|}{Over-fitting} & 11 & 18 & 17\\
 \end{tabular}
 \caption{Classification accuracies for the SVM on data sets pre-processed with \algo{CobFC}, \algo{DC-Fringe}, and the baseline approach\label{svm-accuracies}}
 \end{table*}
 
  Table \ref{nb-accuracies} shows the results for \algo{Naive Bayes}. Obviously, \algo{Naive Bayes}, a classifier challenged when trying to learn concepts that are not linearly separable, benefits strongly from a pre-processing step of the data. This is in line with other results from the feature construction literature, as well as theoretical considerations.
 
 \algo{CobFC} helps \algo{Naive Bayes} improve on two-thirds (21/32) of the data sets, compared to 13 for \algo{DC-Fringe} (which uses many more features), and even the baseline leads to improvements in one-third of the cases. \emph{When} \algo{DC-Fringe} leads to improvements, the gain is occasionally much higher than for \algo{CobFC} but this is paid for by a many cases of over-fitting. This phenomenon is even more pronounced for the baseline, which removes all troublesome points from the training data, rewarding simpler hypotheses.
  
\begin{table*}
 \scriptsize
 \centering
 \begin{tabular}{l|c|c|c|c|c}
 & \multicolumn{2}{c|}{Original} & \algo{CobFC} & DC-Fringe & Baseline\\
 Data set & Train & Test & Test & Test & Test\\\hline
 Breast Cancer (Wisc.) & $97.73 \pm 0.47$ & $94.57 \pm 2.84$& $\mathbf{96.00 \pm 2.50}$ & $\underline{94.14 \pm 2.89}$ & $\underline{93.99 \pm 3.35}$ \\ 
Diabetes & $83.41 \pm 2.04$ & $73.57 \pm 4.92$& $\mathbf{75.27 \pm 5.50}$ & $\underline{70.96 \pm 4.42}$ & $\mathbf{74.22 \pm 3.85}$ \\ 
Ecoli & $93.09 \pm 1.26$ & $81.84 \pm 5.38$& $\mathbf{83.65 \pm 6.00}$ & $\underline{78.26 \pm 6.01}$ & $\mathbf{82.74 \pm 6.25}$ \\ 
Glass & $93.10 \pm 1.19$ & $66.88 \pm 11.23$& $\mathbf{67.86 \pm 11.93}$ & $\mathbf{72.01 \pm 10.06}$ & $60.30 \pm 10.40$ \\ 
Heart Statlog & $92.72 \pm 1.53$ & $76.30 \pm 5.58$& $\mathbf{81.48 \pm 6.76}$ & $\underline{74.81 \pm 9.04}$ & $\mathbf{77.78 \pm 6.98}$ \\ 
Ionosphere & $98.42 \pm 0.39$ & $90.02 \pm 4.53$& $88.03 \pm 2.64$ & $\mathbf{90.30 \pm 5.09}$ & $88.33 \pm 5.74$ \\ 
Iris & $98.00 \pm 0.78$ & $94.00 \pm 6.63$& $93.33 \pm 6.29$ & $\mathbf{94.67 \pm 4.22}$ & $\underline{92.00 \pm 8.78}$ \\ 
Liver Disorders & $85.57 \pm 3.32$ & $64.93 \pm 5.96$& $\mathbf{65.83 \pm 7.26}$ & $\underline{63.23 \pm 9.08}$ & $64.37 \pm 7.55$ \\ 
Optdigits & $98.00 \pm 0.10$ & $90.39 \pm 1.01$& $90.09 \pm 1.73$ & $\mathbf{92.05 \pm 0.99}$ & $\underline{90.23 \pm 0.98}$ \\ 
Pendigits & $99.26 \pm 0.06$ & $96.37 \pm 0.75$& $\mathbf{96.42 \pm 0.69}$ & $\mathbf{96.96 \pm 0.72}$ & $\underline{95.53 \pm 0.66}$ \\ 
Spambase & $97.29 \pm 0.22$ & $92.25 \pm 1.68$& $\mathbf{92.94 \pm 0.10}$ & $\mathbf{92.66 \pm 1.24}$ & $\mathbf{92.76 \pm 1.42}$ \\ 
Page Blocks & $98.59 \pm 0.13$ & $97.02 \pm 0.60$& $\mathbf{97.17 \pm 0.61}$ & $\underline{96.37 \pm 0.83}$ & $\underline{95.91 \pm 0.91}$ \\ 
Segment & $99.14 \pm 0.14$ & $96.54 \pm 1.40$& $96.41 \pm 1.51$ & $\mathbf{96.80 \pm 1.39}$ & $\underline{95.97 \pm 1.29}$ \\ 
Segnoise & $99.02 \pm 0.34$ & $94.87 \pm 1.66$& $93.87 \pm 2.01$ & $\underline{94.27 \pm 1.76}$ & $\underline{93.80 \pm 1.48}$ \\ 
Sonar & $98.13 \pm 0.88$ & $71.17 \pm 10.30$& $\mathbf{74.52 \pm 7.89}$ & $\mathbf{78.88 \pm 8.56}$ & $\mathbf{73.52 \pm 7.79}$ \\ 
Spectrometer & $90.56 \pm 0.58$ & $48.96 \pm 4.04$& $48.79 \pm 5.04$ & $\underline{42.74 \pm 3.22}$ & $\underline{48.21 \pm 3.51}$ \\ 
Vehicle & $91.53 \pm 3.29$ & $74.60 \pm 4.24$& $\underline{73.76 \pm 3.20}$ & $\underline{68.32 \pm 4.21}$ & $\underline{70.09 \pm 3.60}$ \\\hline\hline 
Audiology & $91.15 \pm 1.24$ & $77.02 \pm 5.65$& $\underline{75.26 \pm 5.35}$ & $\mathbf{77.91 \pm 8.37}$ & $\underline{73.93 \pm 6.19}$ \\ 
Breast Cancer & $81.89 \pm 1.77$ & $72.09 \pm 7.10$& $\underline{70.96 \pm 5.58}$ & $\underline{63.29 \pm 6.73}$ & $\underline{70.63 \pm 4.64}$ \\ 
Car & $98.97 \pm 0.21$ & $96.24 \pm 1.14$& $96.06 \pm 3.02$ & $\mathbf{99.71 \pm 0.41}$ & $\underline{93.92 \pm 1.77}$ \\ 
Dermatology & $98.03 \pm 0.36$ & $96.19 \pm 3.89$& $\underline{96.19 \pm 3.89}$ & $\underline{93.46 \pm 3.64}$ & $\underline{95.63 \pm 4.16}$ \\ 
Kr-vs-kp (Chess)& $99.59 \pm 0.11$ & $99.37 \pm 0.55$& $99.25 \pm 0.49$ & $\mathbf{99.47 \pm 0.47}$ & $\underline{99.16 \pm 0.61}$ \\ 
Lung Cancer & $90.23 \pm 4.98$ & $35.00 \pm 29.61$& $35.00 \pm 29.61$ & $\mathbf{36.67 \pm 23.31}$ & $\mathbf{35.83 \pm 28.88}$ \\ 
Lymph & $94.30 \pm 1.42$ & $75.57 \pm 13.77$& $\mathbf{76.29 \pm 15.46}$ & $\mathbf{80.29 \pm 11.02}$ & $\mathbf{77.67 \pm 13.38}$ \\ 
Mol. Biol. (Prom.) & $96.96 \pm 0.92$ & $74.73 \pm 19.04$& $\mathbf{75.64 \pm 17.66}$ & $\mathbf{81.09 \pm 7.74}$ & $\underline{50.00 \pm 3.71}$ \\ 
Nursery & $99.83 \pm 0.02$ & $99.44 \pm 0.25$& $\mathbf{99.84 \pm 0.09}$ & $\mathbf{99.98 \pm 0.03}$ & $\underline{96.59 \pm 0.80}$ \\ 
Postop. Patient Data & $73.21 \pm 2.02$ & $68.89 \pm 10.21$& $\underline{68.89 \pm 10.21}$ & $\underline{57.78 \pm 22.10}$ & $66.67 \pm 11.71$ \\ 
Primary Tumor & $62.41 \pm 1.47$ & $40.73 \pm 9.09$& $\mathbf{43.07 \pm 5.09}$ & $\underline{38.35 \pm 7.35}$ & $\underline{36.88 \pm 9.95}$ \\ 
Soybean & $96.88 \pm 0.68$ & $92.83 \pm 3.62$& $\underline{91.80 \pm 4.22}$ & $\underline{91.81 \pm 3.46}$ & $89.75 \pm 4.68$ \\ 
Splice & $98.18 \pm 0.11$ & $94.17 \pm 1.17$& $\underline{94.04 \pm 1.22}$ & $\underline{92.61 \pm 0.81}$ & $\underline{94.08 \pm 0.96}$ \\ 
Tic-tac-toe & $97.68 \pm 0.48$ & $93.21 \pm 4.08$& $89.98 \pm 4.50$ & $\mathbf{97.18 \pm 2.03}$ & $\underline{90.71 \pm 2.32}$ \\ 
Voting & $97.24 \pm 0.31$ & $95.87 \pm 2.36$& $\mathbf{96.32 \pm 2.22}$ & $\underline{94.72 \pm 4.33}$ & $\underline{94.94 \pm 2.82}$ \\ \hline
\multicolumn{3}{c|}{Improvements} & 15 & 16 & 7\\
\multicolumn{3}{c|}{Over-fitting} & 7 & 16 & 20\\
 \end{tabular}
 \caption{Classification accuracies of J48 on data sets pre-processed with \algo{CobFC}, \algo{DC-Fringe}, and the baseline approach, respectively\label{j48-accuracies}}
 \end{table*}
 
Tuning the SVM's parameters takes so much time for the larger data sets (\emph{Nursery}, \emph{Optdigits}, \emph{Pendigits}), especially in combination with the expensive \algo{DC-Fringe} feature construction, that the experiments had not finished by the time of writing. We do not expect that those data sets would have made a difference in the general trend of the results, however.

The SVM is such a strong learner that it benefits only in few cases from feature construction (see Table \ref{svm-accuracies}), no matter the technique. Instead, the addition of attributes seems to allow too fine-grained modeling of the training data, leading to many more cases of over-fitting than for \algo{Naive Bayes}, even though \algo{CobFC} still performs better than \algo{DC-Fringe}.
 
 \begin{table*}
 \scriptsize
 \centering
  \begin{tabular}{l|c|c|c|c|c} 
  &  & \multicolumn{4}{c}{Minimum Support}\\
 Data set & 1\% & 2\% & 3\% & 4\% & 5\% \\\hline
 Breast Cancer & $\underline{70.18 \pm 10.72}$ & $\underline{71.24 \pm 11.52}$ & $\underline{71.24 \pm 11.52}$ & $\underline{71.24 \pm 11.52}$ & $71.24 \pm 11.52$ \\ 
 Lung Cancer & & & $\underline{49.17 \pm 28.45}$ & & $\underline{49.17 \pm 28.45}$ \\\hline 
 Breast Cancer (Wisc.) & $\mathbf{96.71 \pm 1.65}$ & $\mathbf{96.71 \pm 1.65}$ & $\mathbf{96.86 \pm 1.47}$ & $\mathbf{96.86 \pm 1.47}$ & $\mathbf{96.86 \pm 1.47}$ \\ 
 Diabetes & $\underline{74.36 \pm 5.73}$ & $\underline{74.23 \pm 5.59}$ & $\underline{74.36 \pm 5.73}$ & $\underline{74.09 \pm 4.97}$ & $\underline{74.22 \pm 4.91}$ \\ 
 Liver Disorders & $\mathbf{73.09 \pm 8.70}$ & $\mathbf{73.09 \pm 9.14}$ & $\underline{72.80 \pm 9.06}$ & $\underline{72.52 \pm 9.13}$ & $\underline{72.23 \pm 8.60}$ \\ 
 Spambase & $\mathbf{94.24 \pm 1.23}$ & $\mathbf{94.30 \pm 1.04}$ & $\mathbf{94.08 \pm 0.80}$ & $\mathbf{94.30 \pm 0.75}$ & $\mathbf{94.13 \pm 0.79}$ \\ 
 Page Blocks & $\underline{95.83 \pm 0.79}$ & $\underline{95.91 \pm 0.81}$ & $\underline{95.93 \pm 0.80}$ & $\underline{95.91 \pm 0.81}$ & $\mathbf{96.05 \pm 0.66}$ \\ 
 Breast Cancer & $\underline{71.66 \pm 6.34}$ & $\mathbf{73.07 \pm 7.08}$ & $\mathbf{72.71 \pm 7.03}$ & $\mathbf{73.41 \pm 6.28}$ & $\mathbf{73.08 \pm 6.15}$ \\ 
 Dermatology & $\underline{97.55 \pm 2.69}$ & $\underline{97.55 \pm 2.69}$ & $\underline{97.55 \pm 2.69}$ & $\underline{97.55 \pm 2.69}$ & $\underline{97.55 \pm 2.69}$ \\ 
 Lymph & $\underline{81.05 \pm 8.36}$ & $\underline{80.38 \pm 8.71}$ & $\underline{79.05 \pm 9.73}$ & $\underline{80.38 \pm 8.71}$ & $\underline{79.71 \pm 8.99}$ \\ 
 Mol. Biol. (Prom.) & $\underline{90.36 \pm 7.94}$ & $\underline{90.36 \pm 7.94}$ & $\underline{90.36 \pm 7.94}$ & $\underline{90.36 \pm 7.94}$ & $\underline{90.36 \pm 7.94}$ \\ 
 Soybean & $92.82 \pm 3.94$ & $\underline{92.97 \pm 3.65}$ & $\mathbf{93.12 \pm 3.60}$ & $\underline{93.11 \pm 3.47}$ & $\mathbf{93.26 \pm 3.27}$ \\ 
 Splice & $\underline{96.02 \pm 0.94}$ & $\underline{96.02 \pm 0.94}$ & $\underline{96.02 \pm 0.94}$ & $\underline{96.02 \pm 0.94}$ & $\underline{96.02 \pm 0.94}$ \\ \hline
 Vehicle & $\underline{73.76 \pm 3.20}$ & $\underline{73.76 \pm 3.20}$ & $\underline{73.16 \pm 3.71}$ & $\underline{73.28 \pm 3.68}$ & $\underline{73.40 \pm 3.67}$ \\ 
 Audiology & $\underline{75.26 \pm 5.35}$ & $\underline{75.26 \pm 5.35}$ & $\underline{75.26 \pm 5.35}$ & $\underline{75.69 \pm 4.56}$ & $\underline{75.69 \pm 4.56}$ \\ 
Breast Cancer & $\underline{70.96 \pm 5.58}$ & $\underline{67.87 \pm 6.93}$ & $\underline{68.90 \pm 5.19}$ & $\underline{70.28 \pm 7.24}$ & $\underline{70.97 \pm 6.26}$ \\ 
Dermatology & $\underline{96.19 \pm 3.89}$ & $\underline{96.19 \pm 3.89}$ & $\underline{96.19 \pm 3.89}$ & $\underline{96.19 \pm 3.89}$ & $\underline{96.19 \pm 3.89}$ \\ 
Postop. Patient Data & $\underline{68.89 \pm 10.21}$ & $\underline{68.89 \pm 10.21}$ & $\underline{65.56 \pm 15.23}$ & $\underline{65.56 \pm 15.23}$ & $68.89 \pm 10.21$ \\ 
Soybean & $\underline{91.80 \pm 4.22}$ & $\underline{91.80 \pm 4.22}$ & $\underline{91.80 \pm 3.80}$ & $\underline{91.80 \pm 3.80}$ & $\underline{91.80 \pm 3.80}$ \\ 
Splice & $\underline{94.04 \pm 1.22}$ & $\underline{94.04 \pm 1.22}$ & $\underline{94.04 \pm 1.22}$ & $\underline{94.04 \pm 1.22}$ & $\underline{94.04 \pm 1.22}$ \\ 
    \end{tabular}
\caption{Classification accuracies when using overfitting control via minimum support\label{overfitting-control}}
 \end{table*}

Finally, we would expect the decision tree learner to benefit greatly from \algo{DC-Fringe}'s features since it is its model that drives feature construction and as Table \ref{j48-accuracies} shows, this is indeed the case, \algo{DC-Fringe} showing its best performance. It is only marginally better than \algo{CobFC}, however, and we can also observe that on all data sets for which it does not lead to an improvement, \algo{DC-Fringe} leads to over-fitting. Interestingly, this is also the learner that suffers most from removal of class outliers.

For each classifier, there are data sets for which the \algo{CobFC} features overfit, and Table \ref{overfitting-control} shows the effects of controlling this via a minimum support constraint. \algo{Naive Bayes} results are shown in the two top rows, \algo{SVM} in the eleven underneath, and \algo{J48} in the seven at the bottom. The emphasis is the same as before -- {\bf bold} for improvement over unaugmented, \underline{underlined} for over-fitting. As can be seen, such over-fitting control can be beneficial, increasing the number of data sets for which the \algo{SVM} can improve by six. It also shows, however, that \algo{J48} cannot be helped in this manner, and that -- as too often in pattern mining -- there is no clear-cut mininum support threshold.

\section{Summary and Conclusions\label{conclusion}}

In this paper, we have introduced a new general framework for feature construction. We propose to use outlier detection techniques to identify class outliers, and discriminative pattern mining techniques for deriving features from the $k$-neighborhoods surrounding them. 

We have discussed how our framework addresses the four aspects of feature construction identified in \cite{DBLP:conf/ijcai/MatheusR89}, and instantiated our framework for attribute-valued data using \algo{LOF} as outlier detector, and \algo{C4.5-Rules} as discriminative pattern miner.

In the experimental evaluation, we have demonstrated the usefulness of the framework for \algo{Naive Bayes} and \algo{J48}, while an \algo{SVM} learner could not benefit much. In addition, we have shown that it is far less prone to over-fitting than feature construction that re-iterates on the training data, or removing class outliers from the data. 

Combined with the cheaper pre-processing cost compared to iterative feature construction, and the independence from a particular classifier, we consider this to be evidence for the promise of our new framework.

Since the experimental evaluation used only a proof-of-concept instantiation of the framework, however, there is need for further evaluation to identify effective outlier detection/pattern mining combinations. Also, while over-fitting control by minimum support showed some beneficial effects, the difficulty of choosing a good threshold means that alternative methods would be preferable. We will work on improving this aspect of the framework. Finally, we intend to explore the use of our framework for more complex representations, i.e. structured data.


\bibliography{../bibliographie}

\end{document}